\documentclass[11pt,a4paper]{article}
\usepackage[hyperref]{naaclhlt2019}
\usepackage{times}
\usepackage{latexsym}

\usepackage[shortlabels]{enumitem}
\usepackage[font={footnotesize}]{caption}
\usepackage{graphicx}
\usepackage{amsmath}
\usepackage{amssymb}
\usepackage{xcolor}
\usepackage{algorithm}
\usepackage[noend]{algpseudocode}
\DeclareMathOperator*{\argmax}{arg\,max}

\DeclareMathOperator*{\softmax}{soft\,max}
\usepackage{makecell}
\usepackage{tablefootnote}
\usepackage{subfig}

\usepackage{url}

\newcommand{\vecbf}[1]{\boldsymbol{\mathbf{#1}}}
\newcommand{\sA}{\mathcal{A}}
\newcommand{\sC}{\mathcal{C}}
\newcommand{\sV}{\mathcal{V}}

\newcommand{\bR}{\mathbb{R}}
\newcommand{\ba}{\vecbf{a}}
\newcommand{\bx}{\vecbf{x}}
\newcommand{\bu}{\vecbf{u}}
\newcommand{\bc}{\vecbf{c}}
\newcommand{\cco}{\textsc{CCO}}
\newcommand{\notraincco}{\textsc{NoTrainCCO}}
\newcommand{\pg}{\textsc{PG}}
\newcommand{\obverter}{\textsc{Obverter}}
\newcommand{\pglowent}{\textsc{PGLowEnt}}
\newcommand{\pgfixedparams}{\textsc{PGFixed}}
\newcommand{\fixedrandlang}{\textsc{FixedRand}}

\aclfinalcopy 

\title{Emergence of Communication in an Interactive World with Consistent Speakers}

\author{Ben Bogin, Mor Geva, Jonathan Berant \\
  Tel Aviv University \\
  ben.bogin@cs.tau.ac.il, morgeva@mail.tau.ac.il, joberant@cs.tau.ac.il}

\date{}

\begin{document}
\maketitle
\begin{abstract}
  Training agents to communicate with one another given task-based supervision only has attracted considerable attention recently, due to the growing interest in developing models for human-agent interaction. Prior work on the topic focused on simple environments, where training using policy gradient was feasible despite the non-stationarity of the agents during training. 
In this paper, we present a more challenging environment for testing the emergence of communication from raw pixels, where training using policy gradient fails. We propose a new model and training algorithm, that utilizes the 
structure of a learned representation space to produce more consistent speakers at the initial phases of training, which stabilizes learning. We empirically show that our algorithm substantially improves performance compared to policy gradient. We also propose a new alignment-based metric for measuring context-independence in emerged communication and find our method increases context-independence compared to policy gradient and other competitive baselines.
\end{abstract}

\section{Introduction}


Natural language is learned not by passively processing text, but through active and interactive communication that is grounded in the real world \cite{winograd1972understanding,bnmer1983child,harnad1990symbol}. 
Since grounding is fundamental to human-agent communication, substantial effort has been put into developing grounded language understanding systems, in which an agent is trained to complete a task in an interactive environment given some linguistic input
\cite{siskind1994grounding,roy2002learning,chen2008learning,Wang2016LearningLG,Gauthier2016APF,Hermann2017GroundedLL,Misra2017MappingIA}.

Recently there has been growing interest in developing models for grounded multi-agent communication, where communication arises between neural agents solely based on the necessity to cooperate in order to complete an end task \cite{lewis2017deal,Kottur2017NaturalLD,Lazaridou2018EmergenceOL}. 
Such computational accounts shed light on the properties of the communication that emerges, as a function of various constraints on the agents and environment, and allow us to examine central properties such as compositionality \cite{Nowak1999TheEO,Wagner2003ProgressIT}.
This is an important step towards understanding how to construct agents that develop robust and generalizable communication protocols, which is essential for human-agent communication.

However, most prior work in this field has focused on simple referential games, where a speaker agent communicates a message and a listener agent chooses an answer from a small number of options. This setup suffers from several simplistic assumptions.
First, the listener performs a single action and observes immediate feedback, while in the real world agents must perform long sequences of actions and observe delayed reward. Second, the agents solve a single task, while in the real world agents must perform multiple tasks that partially overlap. This  results in a relatively simple optimization problem, and thus most prior work employed standard policy-gradient methods, such as REINFORCE \cite{Williams1992SimpleSG} to solve the task and learn to communicate.

\begin{figure}[t]
\centering
\includegraphics[scale=0.2]{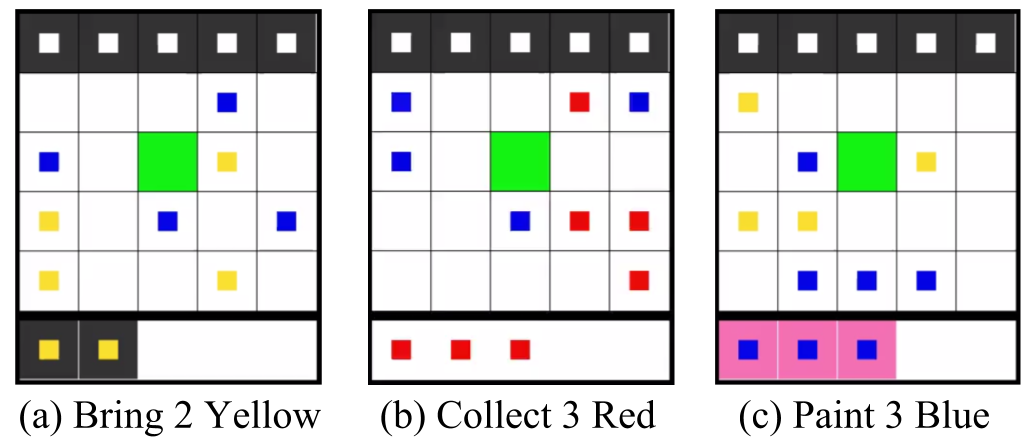}
\caption{Initial worlds as observed by the speaker for the three mission types. The goal inventory is at the bottom and the drop-zone is the top row. In \textsc{Bring}, blocks are marked by black, and in \textsc{Paint}, blocks are marked by pink. The listener is marked by a green square.}
\label{fig:world}
\end{figure}

In this work, we propose a more challenging interactive environment for testing communication emergence, where a speaker and a listener interact in a 2D world and learn to communicate from raw pixel data only. The environment accommodates multiple tasks, such as collecting, using, and moving objects,
where the speaker emits a multi-symbol utterance, and the listener needs to perform a long sequence of actions in order to complete the task. For example, in Figure~\ref{fig:world}a the task is to navigate in the environment, collect exactly two yellow blocks (and no blue blocks) and bring them to a drop-zone.

Training the agents using policy gradient fails in this interactive environment.
This is due to the non-stationarity of both  agents that constantly update their model, the stochasticity of their actions at the initial phase of learning, combined with the long sequence of actions required to solve the task.
In this work, we propose a more stable training algorithm, inspired by the obverter technique \cite{batali1998computational,Choi2018CompositionalOC}, where we impose structure on the learned representation space of utterances and worlds, such that the speaker produces more consistent utterances in similar contexts. This aids the listener in learning to map utterances to action sequences that will solve the task.
We show that the agent's ability to communicate and solve the task substantially improves compared to policy gradient methods, especially as the tasks become increasingly complex.

Once the agents solve a task, we can analyze the properties of the communication protocol that has arisen.
We focus on the property of \emph{context-independence}, namely, whether symbols retain their semantics in various contexts which, under a mild assumption, implies compositionality. To this end, we develop a new alignment-based evaluation metric that aligns concepts with symbols, and find that our algorithm also produces a communication protocol that is more context-independent compared to policy gradient and other competitive baselines.

To summarize, the contribution of this work is three-fold:
\begin{enumerate}[topsep=0pt,itemsep=0ex,parsep=0ex,leftmargin=*]
\item A rich and extensible environment for investigating communication emergence.
\item A new method for training communicating agents that substantially outperforms policy gradient.
\item A novel metric for evaluating context-independence of a communication protocol.
\end{enumerate}

The code-base is available at  	\url{https://github.com/benbogin/
emergence-communication-cco/}.

\section{Related work}
Prior work on emergent communication focused on variants of referential games. In such games, a speaker describes an image or an object, and a listener chooses between what the speaker is referring to and one or more distractors. This setup was introduced in Linguistics and Cognitive Science \cite{lewis1969convention,steels2003,Skyrms2010-SKYSEL,Steels2012TheGN} and has recently been studied from a machine learning perspective
\cite{Lazaridou2016MultiAgentCA,Lazaridou2016TowardsMC,Evtimova2018MultimodalMR,Havrylov2017EmergenceOL,Lazaridou2018EmergenceOL}. In our setup, the agent has to perform a long sequence of actions in an interactive environment. 
Because the space of action sequences is huge and positive feedback is only sparsely received, optimization is a major issue since the probability of randomly performing a correct action sequence is vanishing.
Some work on referential games performed a differentiable relaxation to facilitate end-to-end learning \cite{Choi2018CompositionalOC,Mordatch2017EmergenceOG,Havrylov2017EmergenceOL}, while we in this work use reinforcement learning.

\citet{Mordatch2017EmergenceOG} have shown emergence of communication in an interactive 2D world. 
However, they assume a model-based setup, where the world dynamics are known and differentiable, which limits the applicability of their method.
Moreover, the input to the agents is a structured representation that describes the world and goal rather than raw pixels.

Evaluating the properties of emergent communication is a difficult problem. Prior work has focused on qualitative analyses \cite{Havrylov2017EmergenceOL,Evtimova2018MultimodalMR,Mordatch2017EmergenceOG} 
as well as generalization tests \cite{Kottur2017NaturalLD,Choi2018CompositionalOC}. \citet{Lazaridou2016TowardsMC} have evaluated communication by aligning concepts and symbols, in a single-symbol utterance. Related to our evaluation method,  \newcite{lazaridou2018emergence}, evaluated multi-symbol communication, but did not explicitly test for alignment between symbols and concepts.
In this work we propose an evaluation metric that is based on multi-symbol alignment algorithms, namely IBM model 1 \cite{Brown1993TheMO}.

\section{A Multi-task Grid World} 
\label{sec:world}

In this section we present the environment and tasks.
The environment design was guided by several principles. First, completing the tasks should require a sequence of actions with delayed reward. 
Second, agents should obtain raw visual input rather than a structured representation. Third, multiple related tasks should be executed in the environment, to examine  information transfer between tasks. 

Our environment is designed for a speaker $S$,  and a  listener $L$. The listener observes an initial 2D 5x5 grid world $w_0$, where objects of different colors are randomly placed in various cells of the grid, and an inventory of objects collected so far. The speaker observes $w_0$, but also an image of a \emph{goal inventory} of objects $g$, which specifies the final goal: the type of mission that needs to be performed as well as the number and color of objects that the listener must interact with (Figure~\ref{fig:world}). Given the input pair $(w_0, g)$ the speaker produces an utterance $\bu=(u_1, \dots, u_{|\bu|})$, which includes a sequence of discrete symbols from a vocabulary $\sV$, which can be viewed as the speaker's actions.
The listener generates a sequence of actions $\ba=(a_1, \dots, a_{|\ba|})$. At each step $t$, it observes the 
current world state and utterance 
$(w_t, \bu)$ and selects an action $a_t \in \sA$ from a set of actions, which causes the environment to deterministically modify the world state, i.e., $w_{t+1} = T(w_t, a_t)$. The game ends when the listener either accomplishes or fails the task, or when a maximum number of steps has been reached.

Our environment currently supports three \emph{missions}, which can be viewed as corresponding to different ``verbs'' in natural language:
\begin{enumerate}
[topsep=0pt,itemsep=0ex,parsep=0ex,leftmargin=*]
\item \textsc{Collect} (Figure~\ref{fig:world}b): The agent needs to collect a set of objects. 
When the listener arrives at a position of an object, it automatically collects it. It fails if it passes in positions with objects that should not be collected. After collecting all objects the listener should declare \textsc{Stop}.
\item \textsc{Paint} (Figure~\ref{fig:world}c): The agent needs to collect a set of objects, paint them, and then declare \textsc{Stop}. The action \textsc{Paint} performs a painting action on the object that was collected last. 
\item \textsc{Bring} (Figure~\ref{fig:world}a): The agent needs to collect a set of objects and then bring them to a ``drop-zone''. It fails if it passes in the drop-zone before collecting all necessary objects.
\end{enumerate}

A \emph{task} is defined by a mission, a number of objects and a color.
Note that the 3 missions are overlapping in that they all require understanding of the notion \textsc{Collect}. The set of actions $\sA$ for the listener includes six actions: move to one of the cardinal four directions (\textsc{Up}, \textsc{Down}, \textsc{Left}, \textsc{Right}), 
\textsc{Paint}, and \textsc{Stop}. The actions of the speaker are defined by the symbols in the vocabulary $v \in \sV$, in addition to an END token that terminates the utterance.

The environment provides a per-action return $r_t(\bx, a_t)$ after performing an action $a_t$ in the environment $\bx$. For the speaker, $r_t(\bx, u_{|\bu|}) = 1$ iff the listener successfully completes the task, and is zero otherwise.  That is, the speaker gets a return only after emitting the entire utterance.
For the listener, $r_t(\bx, a_{|\ba|}) = 1$ iff it successfully completed the task.
When $t < |\ba|$, $r_t(\bx, a_t) = -0.1$ to encourage short sequences, and $r_t(\bx, a_t) = 0.1$ for any $t$ where an object is collected, to encourage navigation towards objects.

While the environment is designed for two players it can also be played by a single player.
In that setting, an agent observes both the goal and the current world, and acts based on the observation.
This is useful for pre-training agents to act in the world even without communication.

The speaker outputs an utterance only in the first state ($w_0$), and thus must convey the entire meaning of the task in one-shot, including the mission, color and number of objects. The listener must understand the utterance and perform a sequence of actions to  complete the task. Thus, the probability of randomly completing a task successfully is low, in contrast to referential games where the actor has high probability of succeeding even when it does not necessarily understand the utterance.

\section{Method}

We now describe our method for training the speaker and listener. We first present a typical policy gradient approach for training the agents (A2C, \citet{Mnih2016AsynchronousMF}) and outline its shortcomings. Then, we describe our method, which results in more stable and successful learning.

\subsection{Policy gradient training}
\label{subsec:pg}

As evident from the last section, it is natural to view the dynamics of the speaker and listener as a Markov decision process (MDP), where we treat each agent as if it is acting independently, while observing the other agent as part of the environment $\bx$. Under this assumption, both agents can learn a policy that maximizes expected reward. Thus, policy gradient methods such as REINFORCE \cite{Williams1992SimpleSG} or A2C \cite{Mnih2016AsynchronousMF} can be applied.

For the speaker $S$, the states of the MDP are defined by the input world-goal pair $(w_0, g)$ where the transition function is the identity function (uttered tokens do not modify the world state) and the actions and returns are as described in the previous section.
The goal of the speaker is to learn a policy $\pi_S(w_0, g)$ that maximizes the expected reward:
\begin{equation*}
\mathbb{E}_{\bu \sim \pi_S(w_0, g)}[R(\bx, \bu)],
\end{equation*}
where $\bx$ encapsulates the environment, including the policy of the listener, and $R(\bx, \bu) = \sum_t r_t(\bx, u_t)$. 

For the listener $L$, the states of the MDP at each timestep $t$ are defined by its input world-utterance pair $(w_t, \bu)$, the actions are chosen from $\sA$, and the deterministic transition function is $T(\cdot)$ (defined in last section). 
For a sequence of actions $\ba$ the total reward is $R(\bx, \ba) = \sum_t r_t(\bx, a_t)$, where the returns $r_t$ are as previously defined. Thus, the goal of the listener is to learn a policy $\pi_L$ that maximizes:
\begin{equation*}
\mathbb{E}_{\ba \sim \pi_L(w_0, u)}[R(\bx, \ba)].
\end{equation*}

The expected reward of the speaker and listener is optimized using stochastic gradient descent. The gradient is approximated by sampling $\bu \sim \pi_S(\cdot)$ and $\ba \sim \pi_L(\cdot)$ from multiple games. Because this is a  high-variance estimate of the gradient, we reduce variance by subtracting a baseline from the reward that does not introduce bias \cite{sutton1998reinforcement}. We use ``critic" value functions $v_S(w_0, g)$ and $v_L(w_t, \bu)$ as baselines for the speaker and listener respectively. These value functions predict the expected reward and are trained from the same samples as $\pi_S, \pi_L$ to minimize the L2 Loss between the observed and predicted reward.

A main reason why training with policy gradient fails in our interactive environment, is that the agents are stochastic, and at the initial phase of training the distribution over utterances and action sequences is high-entropy. Moreover, the agents operate in a non-stationary environment, that is, the reward of the speaker depends on the policy of the listener and vice versa. Consequently, given similar worlds observed by the speaker, different utterances will be observed by the listener with high probability at the beginning of training, 
which will make it hard for the listener to learn a mapping from symbols to actions. In referential games this problem is less severe: the length of sequences and size of action space is small, and reward is not sparse. Consequently, agents succeed to converge to a common communication protocol. In our  environment, this is more challenging and optimization fails. We now present an alternative more stable training algorithm.

\subsection{Context-Consistent Obverter}
\label{subsec:cco}

We now describe a new model and training algorithm for the speaker (the listener stays identical to the policy gradient method), which we term \textsc{Context-consistent Obverter (\cco)}, to overcome the aforementioned shortcoming of policy gradient. Specifically, given a task
we would like the speaker to be more consistent, i.e., output the same utterance with high probability even at the initial phase of training, while still having the flexibility to change the meaning of utterances based on the training signal.

Recently, \citet{Choi2018CompositionalOC} proposed a method for training communicating agents. Their work, inspired by the obverter technique  \cite{batali1998computational} which has its roots in the theory of mind \cite{premack1978does}, is based on the assumption that a good way to communicate with a listener is to speak in a way that maximizes the speaker's own understanding.

We observe that under this training paradigm, we can impose structure on the continuous latent representation space that will lead to a more consistent speaker. We will define a model that estimates whether an utterance is good according to the speaker, which will break the symmetry between utterances and lead to a lower-entropy distribution over utterances. This will let the listener observe similar utterances in similar contexts, and slowly learn to act correctly.

Formally, in this setup the speaker $S$ does not learn a mapping $(w_0, g) \mapsto \bu$, but instead learns a  scalar-valued scoring function $\Psi_S(w_0, g, \bu)$ that evaluates how good
the speaker thinks an utterance is, given the initial world and the goal.
Specifically, we define:
\begin{align*}
&\Psi_S(w_0, g, \bu) = -d(h_g, h_u), \\
&h_g = f_g(w_0, g), h_g \in \bR^F, \\
&h_u = f_u(w_0, \bu), h_u \in \bR^F.
\end{align*}
That is, we encode $(w_0, g)$ and $(w_0, \bu)$ as vectors with functions $f_g$ and $f_u$, and score them by the negative euclidean distance between them.\footnote{While we use euclidean distance as $d$, it is not guaranteed that a closer state is better with respect to the actual goal. We empirically found this works well, and leave options such as learning this function for future work.} The best possible encoding of an utterance in this setup is to have the same encoding for $(w_0, g)$ and $(w_0, \bu)$: when $h_u = h_g$, then $\Psi_S = 0$.

\begin{algorithm}[t]
{\footnotesize
    \caption{\cco{} speaker decoder}
    \label{alg:ccodecode}
    \begin{algorithmic}[1]
    \Require world-goal pair $(w_0, g)$
    \State $\bu \leftarrow [] $
    \For{$i=1 \dots n$}
      \State $s \leftarrow \{\} $ // scoring dictionary
      \For{$v$ in $\sV$} $s[v] \leftarrow \Psi_S(w_0, g, [\bu; v])$ \label{line:score}
      \EndFor
      \If {$i>1$} $s[\text{END}] \leftarrow \Psi_S(w_0, g, \bu)$ \label{line:stop}
      \EndIf
      \State $\hat{v} \leftarrow \argmax_{v \in \sV \cup \{\text{END}\}}{s}$ \label{line:argmax}
      \If{$\hat{v} = \text{END}$} break
      \EndIf
      \State append $\hat{v}$ to $\bu$
    \EndFor
    \Return $\bu$
    \end{algorithmic}
    }
\end{algorithm}

We can now score utterances based on the geometric structure of the learned representation space using $\Psi_S(\cdot)$, and decode the highest-scoring utterance $\bu$.
A na\"ive decoding algorithm would be to exhaustively score all possible utterances, but this is inefficient. Instead, we use a greedy procedure that decodes $\bu$ token-by-token (see Algorithm~\ref{alg:ccodecode}). 
At each step, given the decoded utterance prefix, we score all possible continuations (line~\ref{line:score}) including the option of no continuation (line~\ref{line:stop}). Then, we take the one with the highest score (line~\ref{line:argmax}), and stop either when no token is appended or when the maximal number of steps $n$ occurs.\footnote{One could replace $\argmax$ with sampling by forming a probability distribution from the scored utterances with a $\softmax$, however empirically we found that this does not improve results.}

As we explain below, the world-goal representation $h_g$ can be pre-trained and fixed. Thus, the speaker in \cco{} is trained to shift utterance representations $h_u$ closer to $h_g$ when a positive reward is observed, and farther away when a  negative reward is observed.
Specifically, given a decoded utterance $\bu$, we calculate the speaker's advantage value $A=R(\bx, \bu)-v_S(w_0,g)$. The objective is then to maximize the score:
\begin{equation*}
\sum_{\hat{\bu}\in P(\bu)} \alpha \cdot A \cdot \Psi_S(w_0, g, \hat{\bu})  \\
\end{equation*}

where $P(\bu)$ is the set of all prefixes of $\bu$, and the value of $\alpha$ is 1 when $A>0$, otherwise it is $0.5$. Our objective pushes utterances that lead to high reward closer to the world-goal representation they were uttered in, and utterances that lead to low reward farther away.
Since the speaker decodes utterances token-by-token, we consider all prefixes in the objective. Last, the coefficient $\alpha$ puts higher weight on positive reward samples.

\begin{figure}[t]
\centering
\includegraphics[scale=0.3]{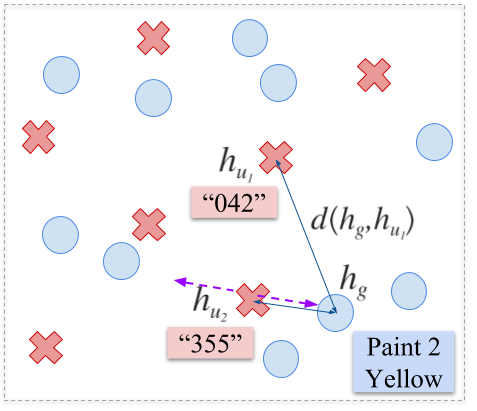}
\caption{An illustration of the learned state space during training. Blue circles are hidden states $h_g$ of input goal, red crosses are hidden states $h_u$ of utterances. The purple dashed arrows indicate that $h_{u_2}$ ($u_2=\text{``355"}$) will become closer to $h_g$ if the advantage $A > 0$, and more distant if $A \leq 0$.}
\label{fig:hidden_space}
\end{figure}

Figure~\ref{fig:hidden_space} illustrates our model and training algorithm. Given the geometric structure of the latent representation space, an utterance with representation $h_{u_2}$ that is close to a certain hidden state $h_g$ will be consistently chosen by the speaker. If this leads to high reward, the model will update parameters such that $h_{u_2}$ becomes closer to $h_g$  and more likely to be chosen for $h_g$ or similar states. Analogously, if this leads to low reward $h_{u_2}$ will become more distant from $h_g$ and eventually will not be chosen. The geometric structure breaks the symmetry between utterances at the beginning of training and stabilizes learning.

\paragraph{Comparison to \citet{Choi2018CompositionalOC}}
Our work is inspired by \citet{Choi2018CompositionalOC} in that the speaker  does not directly generate the utterance $\bu$ but instead receives it as input and has a decoding procedure for outputting $\bu$. However, we differ from their setup in important aspects:
First, we focus on exploiting the structure of the latent space to break the symmetry between utterances and get a consistent speaker, as mentioned above.
Second, their model only works in a setting where the agent receives immediate reward (a single-step referential game), and can thus be trained with maximum likelihood. Our model, conversely, works in an environment with delayed reward. Third, their model only updates the parameters of the listener, and the signal from the environment is not propagated to the speaker. We train the speaker using feedback from the listener. Last, we will show empirically in the next section that our algorithm improves performance and interpretability.

\subsection{Neural network details} \label{neural}

\begin{figure}[t]
\centering
\includegraphics[scale=0.2]{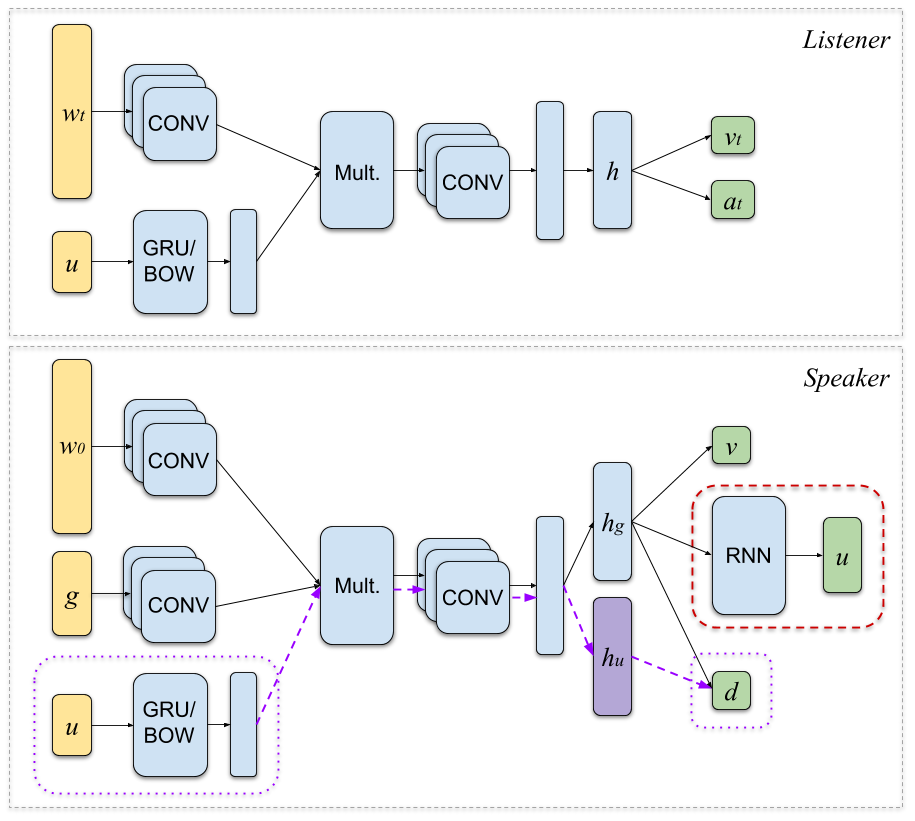}
\caption{A high-level overview of the network architectures of the listener (top) and the speaker (bottom).}
\label{fig:network}
\end{figure}

Figure~\ref{fig:network} provides a high-level overview of the listener and speaker network architectures. 

The listener receives the world-utterance pair ($w_t$, $\bu$). The utterance is encoded using either a GRU \cite{cho2014properties} or a bag-of-words average of embeddings, while the visual input is encoded with convolutional layers. The two encodings are multiplied \cite{oh2015action} and go through convolutional and feed-forward layers to output an action $a_t$ and an estimated critic value $v_t$.

The speaker receives as input the world-goal pair $(w_0,g)$ and encodes it in a similar way to obtain a hidden representation $h_g$ for the input task, which is used to output $v$. When training with policy gradient (marked as a dashed red box), the output utterance $\bu$ is generated with a GRU that receives $h_g$ as its initial hidden state. When training with \cco{}, $\bu$ is given as a third input to the network (marked as a dotted purple box), and $h_u$ is encoded without using $g$ (effectively identical to the listener's architecture). The distance $d$ between $h_u$ and $h_g$ can then be calculated.

\subsection{Pre-training}
\label{subsec:pretraining}
Since we are interested in the emergence of communication, we allow both the listener and the speaker to pre-train and learn to solve all tasks in a single-agent setting.
The speaker and the listener are pre-trained separately using the policy-gradient speaker network, observing only the world and goal and outputting $v$ and $a$. After pre-training, the listener learns to represent $\bu$ from scratch, since it only observed $g$ at pre-training time. The speaker only updates the parameters that represent $\bu$ (dashed purple box in Figure~\ref{fig:network}).

\section{Experimental Evaluation}
\label{sec:experiments}

\begin{table*}[t]
{\scriptsize
\begin{center}
\begin{tabular}{l|l|l|l|l|l|l}
\bf Training setup & \bf Ref-8C/5S & \bf 3C/1N/1M & \bf 8C/1N/1M & \bf 8C/3N/1M & \bf 3C/3N/2M & \bf 3C/3N/3M \\ 
\textbf{V. size / max len.} & 15/20 & 15/20 & 15/20 & 15/20 & 15/20 & 15/20 \\
\hline
\pg & $0.91\pm0.04$ & $0.91\pm0.03$ & $0.78\pm0.05$ & $0.26\pm0.07$ & $0.13\pm0.08$ & $0.05\pm0.01$ \\ 
\pglowent & $0.74\pm0.07$ & $0.92\pm0.02$ & $0.90\pm0.05$ & $0.42\pm0.00$ & $0.31\pm0.03$ & $0.22\pm0.04$ \\
\pgfixedparams & $0.95\pm0.03$ & $0.83\pm0.03$ & $0.70\pm0.00$ & $0.22\pm0.08$ & $0.11\pm0.06$ & $0.05\pm0.00$ \\
\notraincco & $0.85\pm0.05$ & $0.96\pm0.02$ & $0.83\pm0.03$ & $0.28\pm0.04$ & $0.26\pm0.05$ & $0.14\pm0.09$ \\
\fixedrandlang & $1.00\pm0.00$ & $1.00\pm0.00$ & $1.00\pm0.00$ & $1.00\pm0.01$ & $0.35\pm0.00$ & $0.22\pm0.01$ \\
\obverter - GRU & $0.97\pm0.00$ & - & - & - & - & - \\
\obverter - BOW & $0.98\pm0.01$ & - & - & - & - & - \\
\cco - GRU & $1.00\pm0.00$ & $1.00\pm0.00$ & $1.00\pm0.00$ & $0.99\pm0.01$ & $0.90\pm0.07$ & $0.85\pm0.05$ \\
\cco - BOW & $0.99\pm0.00$ & $1.00\pm0.00$ & $1.00\pm0.00$ & $1.00\pm0.01$ & $0.95\pm0.08$ & $0.93\pm0.02$ \\
\end{tabular}
\end{center}
\caption{Test results of task completion performance for tasks with increasing number of colors, numbers and missions (in the referential game we also use shapes). We also provide the vocabulary size and maximum sentence length for each setup. }
\label{tab:perf}
}
\end{table*}

\subsection{Evaluation of Task Completion}

First, we evaluate the ability of the agents to solve tasks given different training algorithms.
We evaluate agents by the proportion of tasks they solve on a test set of 500 unseen worlds. We run each experiment 3 times with different random seeds. The following baselines are evaluated:
\begin{itemize}[topsep=0pt,itemsep=0ex,parsep=0ex,leftmargin=*]
  \item \cco{}-GRU: Our main algorithm.
  \item \cco{}-BOW: Identical to \cco{}-GRU except we replace the GRU that processes $\bu$ with a bag-of-words for both the speaker and the listener.
  \item \obverter: A reimplementation of the obverter as described in \citet{Choi2018CompositionalOC}, where the speaker and listener are first pre-trained separately to classify the color and shape of given objects. The convolution layers parameters are then frozen, similar to CCO.
  \item \pg: A2C policy gradient baseline \cite{Mnih2016AsynchronousMF} (as previously described).
  \item \pglowent: Identical to \pg{}, except that logits are divided by a temperature ($=0.3$) to create a low-entropy distribution, creating a more consistent policy gradient agent.
  \item \pgfixedparams{}: Identical to \pg{}, except that for the speaker only the parameters of the RNN that generates $\bu$ are trained, similar to \cco{}.
  \item \notraincco: Identical to \cco{} but without training the speaker, which results in a consistent  language being output by the decoder.
  \item \fixedrandlang{}: An oracle speaker that given the task (\emph{``paint 1 blue"}) assigns a fixed, unambiguous, but random sequence of symbols. This results in a perfectly consistent communication protocol that does not emerge and is not compositional.
\end{itemize}

We evaluate on the following tasks:

\noindent
\textbf{Referential}: A referential game re-implemented exactly as in \citet{Choi2018CompositionalOC}. The speaker sees an image of a 3D object, defined by one of 8 colors and 5 shapes and describes it to the listener. The listener sees a different image and has to determine if it has the same color and shape.

\begin{figure}[t]
\vspace*{-20pt}
\centering
\includegraphics[scale=0.55]{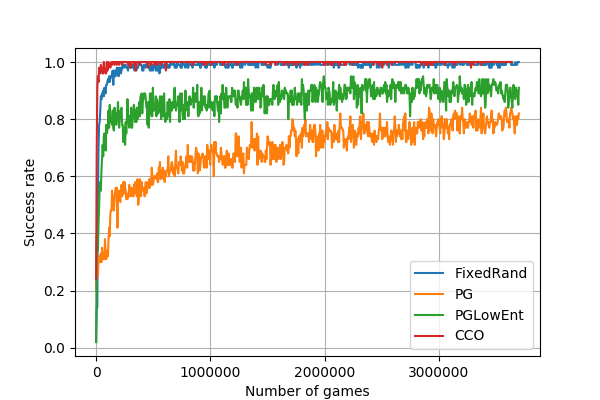}
\caption{Success rate in the 8C/1N/1M task as a function of games played, for vocabulary size 15 and maximum sentence length 20, for  different algorithms.}
\label{fig:performance}
\end{figure}

\noindent
\textbf{Interactive:} 
Our environment with varying numbers of colors, numbers and missions. In Table~\ref{tab:perf}, 3C/3N/3M corresponds to a world with three colors, three numbers and three missions. If a single number is used, it is $1$; If a single mission is used, it is \textsc{Collect}; if two missions are used, they are \textsc{Collect} and \textsc{Paint}.

Table \ref{tab:perf} shows the success rate and standard deviation for each experiment. As the difficulty of the tasks increases, the performance of \pg{} substantially decreases.
\pglowent{}, which is more consistent, outperforms \pg{} but fails on complex tasks. \pgfixedparams{}, which only trains the utterance generation parameters, does not perform better.

\notraincco{} does not perform well, showing that the speaker's learning is essential. 
The consistent and unambiguous oracle \fixedrandlang{} outperforms other baselines, showing that non-stationarity is a core challenge for \pg{}, and a random but consistent language improves learning. However, with more complex tasks the listener is no longer able to decipher the random language, and performance drops.\footnote{We compared networks with similar capacity and training time. Naturally, given a larger model and more training time the listener is likely to memorize a random language.} 
The \obverter{} algorithm solves the referential task almost perfectly, but cannot be used in interactive settings. Finally, our algorithm, \cco{}, performs well on all tasks.

Figure \ref{fig:performance} provides a learning curve for the success rate of different algorithms. The consistent methods \cco{} and \fixedrandlang{} solve the task much faster than \pg{} methods, with a slight advantage for \cco.

\subsection{Evaluation of Language}

We analyze the properties of the emerged language. A hallmark property of natural language is compositionality: the meaning of the whole is a function of the meaning of its parts \cite{Frege1892uber}. However, there is no standard metric for evaluating compositionality in emerged communication. We therefore propose to measure \emph{context-independence}: whether atomic symbols retain their meaning regardless of context. If communication is perfectly context-independent, and under the mild assumption that there exist multiple words in the language that refer to multiple properties and can be combined in a single utterance (which happens in our task and in natural language), then the meaning of the whole is compositional. Natural language is not context-independent, but words often retain  semantics in many contexts. 

We propose a metric that measures to what extent there is a one-to-one alignment between task concepts and utterance symbols.
The measure should provide a high score iff each concept (\textsc{Red}, or \textsc{Paint}) is mapped to a single symbol ("7"), and that symbol is not mapped to other concepts. We base our measure on probabilities of vocabulary symbols given concepts $p_{\text{vc}}(v \mid c)$ and concepts given vocabulary symbols $p_\text{cv}(c \mid v)$, estimated by IBM model 1 \cite{Brown1993TheMO}. We use an IBM model, since it estimates alignment probabilities assuming a hard alignment between  concepts and symbols, and specifically IBM model 1, because the order of concepts is not meaningful.

To run IBM model 1, we generate 1,000 episodes for each model, from which we produce pairs $(\bu, \bc)$ of utterances and task concepts. We run IBM model 1 in both directions, which provides the probabilities $p_\text{cv}(c \mid v)$ and $p_\text{vc}(v \mid c)$.
We now define context-independence score:
\begin{align*}
\forall c: \quad  v^c &= \argmax_v p_{cv}(c \mid v),\\
CI(p_{cv}, p_{vc}) &=\frac{1}{|\sC|} \sum_c p_{vc}(v^c \mid c) \cdot  p_{cv}(c \mid v^c),
\end{align*}
where $\sC$ is the set of all possible task concepts.
The score is the average alignment score for each concept, where the score for each concept is a product of probabilities $p_{vc}$ and $p_{cv}$ for the symbol $v$ that maximizes $p_{cv}(c \mid v)$. This captures the intuition that each concept should be aligned to one vocabulary symbol, and that symbol should not be aligned to any other concept. The measure is one-sided because usually $|\sV| > |\sC|$ and some vocabulary symbols do not align to any concept. Also note that $CI(\cdot)$ will be affected by $|V|$ and $|C|$ and thus 
a fair comparison is between setups where the size of vocabulary and number of concepts is identical. We note that a perfectly context-independent language would yield a score of 1.

\begin{table}[t]
\scriptsize
\begin{center}
\begin{tabular}{l|l|l|l}
\bf Training setup & \bf 3C/3N/1M & \bf 5C/3N/1M & \bf 8C/3N/1M \\ 
\textbf{V. size / max length} & 8/20 & 10/20 & 13/20 \\
\hline
random speaker & $0.03\pm0.00$ & $0.02\pm0.00$ & $0.01\pm0.00$ \\
\fixedrandlang & $0.17\pm0.06$ & $0.12\pm0.03$ & $0.09\pm0.02$ \\
\cco - GRU & $0.36\pm0.10$ & $0.50\pm0.11$ & $0.29\pm0.07$ \\
\cco - BOW & $0.34\pm0.10$ & $0.62\pm0.10$ & $0.30\pm0.08$ \\
\hline
\hline
\\
\bf Training setup &  \bf Ref-8C/5S & \bf 3C/1N/1M & \bf 8C/1N/1M \\ 
\textbf{V. size / max length} & 15/20 & 5/20 & 10/20 \\
\hline
\pg & $0.03\pm0.01$ & $0.20\pm0.05$ & $0.08\pm0.02$ \\
\obverter - GRU & $0.11\pm0.08$ & - & - \\
\obverter - BOW & $0.19\pm0.10$ & - & - \\
\cco - GRU & $0.27\pm0.06$ & $0.45\pm0.03$ & $0.37\pm0.11$ \\
\cco - BOW & $0.40\pm0.02$ & $0.54\pm0.16$ & $0.44\pm0.05$ \\
\hline
\hline
\\
\bf Training setup & \bf 3C/3N/1M & \bf 3C/3N/2M & \bf 3C/3N/3M \\ 
\textbf{V. size / max length} & 11/20 & 11/20 & 11/20 \\
\hline
\cco - GRU & $0.42\pm0.2$ & $0.28\pm0.10$ & $0.20\pm0.02$ \\
\cco - BOW & $0.29\pm0.05$ & $0.29\pm0.05$ & $0.25\pm0.02$ \\
\end{tabular}
\end{center}
\caption{Context-independence evaluation on the test set.}
\label{tab:language}
\end{table}

Table~\ref{tab:language} shows evaluation results, where we run each experiment 3 times. Table~\ref{tab:language}, top, compares \cco{} to the following variants on the mission \textsc{Collect}. A random speaker that samples an utterance randomly sets a lower-bound on $CI(\cdot)$ for reference, and indeed values are close to 0. \fixedrandlang{} is a consistent speaker that is unambiguous, but is not compositional. Unambiguity increases $CI(\cdot)$ slightly, but still the score is low. The $CI(\cdot)$ score of \cco{} is higher, but when we replace the GRU that processes $\bu$ with a bag-of-words, $CI(\cdot)$ improves substantially.

Table~\ref{tab:language}, middle, compares \cco{} to \obverter{} and \pg{}. In the referential game, language emerged with \cco{} is substantially more context independent than \obverter{} and \pg, especially with the BOW variant. \pg{} exhibits very low context-independence in both the referential game and when tested on \textsc{Collect} with varying colors. In contrast to \cco, \pg{} encodes the message in much longer messages, which might explain the low $CI(\cdot)$ score.


Table~\ref{tab:language}, bottom, investigates a multi-task setup, to examine whether training on multiple tasks improves context-independence. We run the agents with 3 colors and 3 numbers, while increasing the number of missions. For a fair $CI(\cdot)$ comparison, we calculated the score using only the colors and numbers concepts. Despite our prior expectation, we did not observe improvement in context-independence as the number of missions increases.

\begin{table}
\scriptsize
\begin{center}
\begin{tabular}{l|l|l|l|l|l}
\bf & \textsc{Yellow} & \textsc{Red} & \textsc{Maroon} & \textsc{Gray} & \textsc{Blue} \\ 
\hline
ONE & 7,6 & 2,6 & 4,6 & 5,6 & 8,6 \\
TWO & 7,0 & 2,0 & 4,0 & 5,0 & 8 \\
THREE & 7,3 & 2,3 & 4,3 & 1 & 9 \\
\end{tabular}
\end{center}
\caption{An example mapping of tasks to utterances showing the most commonly used utterances for the \textsc{\cco{} - BOW} model, in a setting of 5 colors, 3 numbers and 1 mission with a vocabulary size of 10 and a maximum sentence length of 20. The above language results in a $CI(\cdot)$ score of 0.74. }
\label{tab:sample-language}
\end{table}

Table \ref{tab:sample-language} shows a mapping from tasks to utterances, denoting the most frequently used utterance for a set of \textsc{Collect} tasks. The table illustrates a relatively context-independent language ($CI(\cdot) = 0.74$). Some alignments are perfect (symbol `7' for \textsc{YELLOW}), while some symbols are aligned well to a concept in one direction, but not in the other (the symbol `8' always refers to \textsc{Blue}, but \textsc{Blue} isn't always described by `8').

\section{Conclusions}
This paper presents a new environment for testing the emergence of communication in an interactive world. We find that policy gradient methods fail to train, and propose the \cco{} model, which utilizes the structure of the learned representation space to develop a more consistent speaker, which stabilizes learning. We also propose a novel alignment-based evaluation metric for measuring the degree of context-independence in the emerged communication. We find that \cco{} substantially improves performance compared to policy gradient, and also produces more context-independent communication compared to competitive baselines.

\section*{Acknowledgments}
This work was supported in part by the Yandex Initiative in Machine Learning.

\bibliography{naaclhlt2019}
\bibliographystyle{acl_natbib}



\end{document}